\def\year{2022}\relax
\documentclass[letterpaper]{article} 
\usepackage{aaai22}  
\usepackage{times}  
\usepackage{helvet} 
\usepackage{courier}  
\usepackage[hyphens]{url}  
\usepackage{graphicx} 
\usepackage{comment}
\usepackage{color}
\usepackage{subfigure}
\usepackage{natbib}  
\usepackage{caption} 
\newcommand{\tabincell}[2]{\begin{tabular}{@{}#1@{}}#2\end{tabular}}

\title{From Consensus to Disagreement: Multi-Teacher Distillation \\ for Semi-Supervised Relation Extraction}
\author{
    Wanli Li\textsuperscript{\rm 1},
    Tieyun Qian\textsuperscript{\rm 1}\thanks{Corresponding authors.}
    \\
}
\affiliations{
    \textsuperscript{\rm 1}Wuhan University, China\\

    wanli.li@whu.edu.cn, qty@whu.edu.cn

}
\begin{document}

\maketitle

\begin{abstract}
Lack of labeled data is a main obstacle in relation extraction. Semi-supervised relation extraction (SSRE) has been proven to be a promising way for this problem through  annotating unlabeled samples as additional training data. Almost all prior researches along this line adopt multiple models to make the annotations more reliable by taking the intersection set of predicted results from these models. However, the difference set, which contains rich information about unlabeled data, has been long neglected by prior studies.

In this paper, we propose to \emph{learn not only from the consensus but also the disagreement} among different models in SSRE. To this end, we develop \emph{a simple and general multi-teacher distillation (MTD) framework}, which can be easily integrated into any existing SSRE methods. Specifically, we first let the teachers correspond to the multiple models and select the samples in the intersection set of the last iteration in SSRE methods to augment labeled data as usual. We then transfer the class distributions for samples in the difference set as soft labels to guide the student. We finally perform prediction using the trained student model.  Experimental results on two public datasets demonstrate that our framework significantly promotes the performance of the base SSRE methods with pretty low computational cost.


\end{abstract}

\section{Introduction}
The target of relation extraction (RE) is to detect the semantic relation between/among the entities mentioned in the text.
For example,  given a sentence ``\emph{The implant is placed into the jaw bone}'' and two entities \emph{`e1: implant'} and \emph{`e2: jaw bone'},
we aim to identify the \emph{`entity:destination (e1, e2)'} relation.
RE can improve the access and management of text information~\cite{relation_survey}. It is a fundamental task in many natural language processing applications like knowledge graph and question answering.


\begin{figure}[t]
\center{
\includegraphics[scale=0.3]{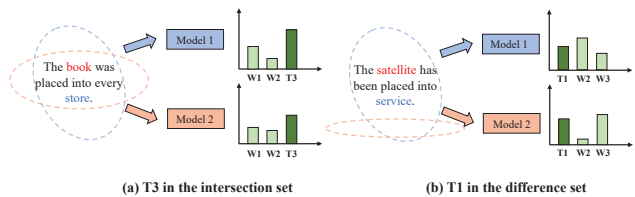}}
\caption{Illustration of the intersection and difference set of the predictions made by two models in SSRE. $T$ and $W$ denote the true and wrong predictions.}
\label{fig:diff}
\vspace{-0.5cm}
\end{figure}

Existing RE methods can be classified into three categories: supervised, distant supervision, and semi-supervised.
Early studies on supervised RE mostly adopt feature-based~\cite{feature_based_1,feature_based_2,feature_based_3} and kernel-based
methods~\cite{kernel_based_1,kernel_based_2,kernel_based_3,kernel_based_4}. Recently deep learning methods~\cite{PCNN,prnn,AGGCN,bert4re,PengGHLLLSZ20} become the mainstream.
The drawback of  supervised RE is that it requires  sufficient hand-labeled data for model training.

To alleviate the expensive human-annotation issue in supervised RE, several distant supervision methods~\cite{distant_supervision,PCNN,shifted_label_distribution,AAAI21_distant} have been proposed for annotating sentences in an unlabeled corpus  with the help of external knowledge bases (KBs). However, the sparse matching results and context-agnostic label noises ~\cite{MetaSelfTraining20} make it hard to apply distant supervision for label generation.

Another line of research is semi-supervised relation extraction (SSRE)~\cite{lp_1,self-training}, which starts from  a small number of labeled data and then gradually augments the labeled data by iteratively training the model and using the model to annotate samples in unlabeled data. SSRE suffers from \emph{the semantic drift problem} ~\cite{semantic_drift07,DualRE}, i.e., the predictions on the unlabeled data might be incorrect. The problem will be   extremely serious when the number of labeled data is limited.

To overcome the semantic drift  problem, conventional SSRE methods often employ multiple models~\cite{self-ensembling,mean-teacher,DualRE}, which
use different seeds for initialization,  different parameter spaces for training, or different modules to complement each other.
The intuition is that multiple models provide  different perspectives on the labeled data, and the semantic drift problem can be partially solved via the collaboration of these models.

The multiple models in existing SSRE methods collaborate by taking the intersection set of their prediction results to generate high-quality labels, as shown in Fig.~\ref{fig:diff} (a). Note that the top-ranked score is used to assign \emph{the hard label} to the instance. That is to say, only both classifiers have the same hard label on a specific instance, can this label be qualified for tagging the instance.
While obtaining more reliable labels, current SSRE methods discard the difference set of prediction results,  which incurs \emph{a big information loss}. On one hand, the results in the difference set made by multiple classifiers may still contain correct predictions.
On the other hand, even if none of the results in this set is correct, the predictions convey intra and inter-class distribution information, as shown in Fig.~\ref{fig:diff} (b). For example, though two models both make wrong predictions (W3 and W2), their secondly ranked prediction scores are in the same class. In such a case, the prediction T1 is correct.

In light of this, we decide to exploit the  knowledge distillation (KD) technique~\cite{Bucilua_ModelCompression_kdd06,Ba_Deep_NIPS14,distillation_hinton} to transfer the prediction knowledge in the difference set. KD is originally proposed to transfer the information from a large teacher model to a small student model. It has been successfully applied to many machine learning fields.
In this paper, we develop \emph{a simple and general KD model for semi-supervised relation extraction}. Specifically, we first adopt multiple models in the final step of existing SSRE methods as multiple teacher models and augment the labeled data using the samples in the  intersection set of these teachers. We then transfer the prediction probabilities for samples in the difference set made by multiple teachers on the augmented training data,  in the form of soft labels, to the student network. We finally train the student with both the augmented data and the soft labels. The trained student model is used for testing.


To the best of our knowledge, there is only one reference~\cite{KD4NRE} exploring KD for RE tasks. However, that method is designed for supervised RE, and it relies on the entire corpus to excavate type constraints as soft rules. Under the semi-supervised circumstance, its performance deteriorates dramatically, as we will see in our experiments.
In summary, our main contributions are as follows.
\begin{itemize}
\item We propose to utilize not only the intersection set but also the difference set of multiple models, which frames semi-supervised learning within a new paradigm.
\item We develop a knowledge distillation framework by converting multiple models into teachers and transferring their disagreements as soft labels to the student.
\item Extensive experimental results demonstrate that our framework can significantly improve the performance of original semi-supervised models.
\end{itemize}

\section{Preliminary}
We first present the problem definition and then give a brief introduction to knowledge distillation.

\subsection{Problem Definition}
\noindent\textbf{Definition 1}  (\emph{Relation Extraction (RE)})
Let \textit{d} = [$t_{1}$, ..., $t_{m}$] be a sentence  with $m$ tokens, and \textit{e$_1$} and \textit{e$_2$} be two entity mentions in \textit{d}. $R$ =\{$r_{1}$, ..., $r_{|R|}$\} is a predefined relation set.
The relation extraction~\footnote{There are three types of relation extraction tasks: sentence-level RE, cross-sentence n-ary RE, and document-level RE. Since prior semi-supervised relation extraction methods are conducted on sentence-level tasks, we also adopt this setting.} task is formulated as a classification problem that determines whether a relation $r$ $\in$ $R$ holds for \textit{e$_1$} and  \textit{e$_2$}.


\noindent\textbf{Definition 2} (\emph{Semi-supervised relation extraction (SSRE)}) Given a set of labeled and unlabeled relation mentions $D_L=\{(d_i,r_i)\}_{i=1}^{|L|}$ and $D_U=\{(d_i)\}_{i=1}^{|U|}$, respectively, the goal of SSRE task is to train a model that fits the labeled data $D_L$, and captures the information in the unlabeled data $D_U$ for augmenting the labeled data. The trained model is used to predict the relation of the samples in the unseen test data $D_T=\{(d_i)\}_{i=1}^{|T|}$.

\subsection{Knowledge Distillation}
Knowledge distillation is originally proposed for model compressing~\cite{Bucilua_ModelCompression_kdd06,Ba_Deep_NIPS14,distillation_hinton},
with the basic idea of transferring the knowledge from the large teacher model $T$ to the small student model $S$.
The obvious way to transfer the knowledge is to use the class probabilities produced by the teacher model as ``\emph{soft labels}'' for training the student  model.

The teacher network generates \emph{the softened class probabilities} with a converting operation, i.e., raising the temperature of the final softmax layer until the teacher produces a suitably soft set of targets~\cite{distillation_hinton}.
\begin{equation}
\small
\label{teacher_softmax}
\widetilde{P}_T = softmax(\widetilde{Z}_T/\tau),
\end{equation}
where $\widetilde{Z}_T$ is the logits produced by the teacher network. $\tau$ is the temperature and  a higher value for $\tau$ produces a softer probability distribution over classes.

The student has the same neural architecture as the teacher. The class probabilities produced by the student model $\widetilde{P}_S$ is the
output of the softmax layer:
\begin{equation}
\small
\label{student_softmax}
\widetilde{P}_S = softmax(\widetilde{Z}_S),
\end{equation}
where $\widetilde{Z}_S$ is the logits produced by the student network.

The student model is trained on labeled data to get the hard labels, and the transferred soft labels are used for distillation supervision. The distillation optimization function $\mathcal{L}_{KD}$ is defined as:
\begin{equation}
\small
\label{L_kd}
\mathcal{L}_{KD} = KL(\widetilde{P}_T||\widetilde{P}_S) = \sum_i^{|D_L|}\widetilde{P}_T(i)log(\widetilde{P}_T(i)/\widetilde{P}_S(i)),
\end{equation}
where $KL$ denotes the Kullback-Leibler divergence, $\widetilde{P}_T(i)$ and $\widetilde{P}_S(i)$ denote the $i$-th element in $\widetilde{P}_T$ and $\widetilde{P}_S$, respectively.

\section{Methodology}
\subsection{Model Overview}
We present an overview of our proposed multiple teacher distillation (MTD) framework for SSRE.
The biggest challenge in SSRE is the lack of labeled data, which leads to a low-quality classifier and the severe semantic drift problem. To overcome this problem, existing SSRE methods~\cite{mean-teacher,DualRE,MetaSelfTraining20} often train several different classifiers on the same data and get the intersection set of their predictions on unlabeled data for data augmentation. This is a simple and effective strategy. While benefiting from the intersection set, current SSRE methods discard the difference set of the predictions from multiple models. As illustrated in the introduction part, this will incur a big information loss. Hence we propose our multiple teacher distillation (MTD) framework.

\begin{figure}[t]
\center{
\includegraphics[scale=0.42]{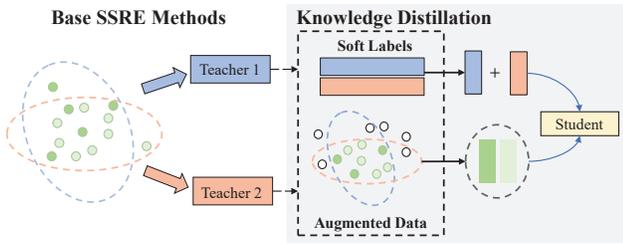}}
\vspace{-0.3cm}
\caption{An overview of our MTD framework for SSRE.}
\label{fig:overview}
\vspace{-0.6cm}
\end{figure}

An overview of our  framework is shown in Fig.~\ref{fig:overview}. MTD is built upon existing SSRE methods. It takes the multiple (usually two) models in SSRE methods as multiple teachers. It then conducts knowledge distillation. To be specific, the multiple teachers' predictions generate \emph{an intersection set} and \emph{a difference set}, and the samples in the intersection set are selected to augment the training data. Meanwhile, the class distributions for samples in the difference set are used to produce soft labels for knowledge distillation. The student model is trained on augmented data with the soft labels.

\vspace{-0.1cm}
\subsection{Encoder}
Before training the models, it is necessary to encode the tokens in input sentences into latent vectors. A variety of encoders have been proposed for RE tasks, such as convolutional neural network (CNN)~\cite{PCNN}, recurrent neural network (RNN)~\cite{prnn}, graph based neural network (GNN)~\cite{AGGCN,KD4NRE}, and pre-trained language model (PLM) encoders~\cite{bert4re,YamadaASTM20}. Among them, PLM encoders can learn universal language knowledge from the large corpus, and are beneficial for downstream tasks.

In view of this, we follow the recent practice~\cite{bert4re} and use BERT~\cite{DevlinCLT19} as PLM for all SSRE methods.  In particular, we insert a special token `[E1]' and `[E2]' at the beginning and the end of the first and the second entity, respectively. We also add a `[CLS]'  and a `[SEP]' token at the beginning and the end of the sentence. Given the running example sentence, its input to the encoder is as follows.

`\textit{[CLS] The [E1] implant [E1] is placed into the [E2] jaw bone [E2] [SEP].}

\subsection{Multiple Teacher Models}
\noindent\textbf{Multiple Models in Existing SSRE Methods} Almost all existing SSRE methods contain two models to obtain more reliable results. For example, Mean-Teacher~\cite{mean-teacher} and Re-Ensemble~\cite{self-ensembling} train two models using different dropout rates and different seeds. DualRE~\cite{DualRE} introduces a retrieval module and a prediction module to improve each other.

\noindent\textbf{Converting Multiple Models into Multiple Teachers} To be compatible with existing SSRE methods (termed as \emph{base methods} hereafter), we make the least change to these methods and directly take the multiple models from the last iteration when training the base methods as the multiple teachers in our MTD framework. In case that the base method Self-Training~\cite{self-training} contains one model, we simply run it two times with different seeds to get two teacher models.

\subsection{Collaboration among Multiple Teachers}
\noindent\textbf{Labeled Data Augmentation}
To obtain high-confidence samples, SSRE methods choose the samples in \emph{the intersection set} of predictions by two models to augment the training data. In our MTD framework, two teacher models are also used to classify the samples in the unlabeled data. If two teachers reach a consensus prediction on a sample, this sample will be added to the training data.
The augmented training data is termed as the ``teaching data'' $\mathcal{D}_{A}$:
\begin{equation}
\label{L_DT}
\mathcal{D}_{A} = \mathcal{D}_L \cup \mathcal{D}_{I},
\end{equation}
where $\mathcal{D}_L$ is the original labeled data, and $\mathcal{D}_{I}$ is composed of instances on which two teachers assign the same label.

\noindent\textbf{Distilling Knowledge}
The way of utilizing the intersection set in our MTD framework is consistent with that in base SSRE methods. However, we move one step further by leveraging \emph{the difference set} of predictions beyond the intersection set. The rationale is that the predictions reflect  the judgments by two teachers and contain  rich intra- and inter-class distribution information. For example, the relationship between ``per:city\_of\_death'' and ``per:country\_of\_death'' is closer than that between  ``per:city\_of\_death'' and ``org:founded\_by''.

To this end, we propose to exploit the power of knowledge distillation in transferring knowledge from the teacher to the student, where the knowledge here is just the class distribution of samples in the difference set generated by two teachers on $\mathcal{D}_A$:
\begin{equation}
\small
\label{teacher_softmax}
\widetilde{P}^m_T = softmax(\widetilde{Z}^m_T/\tau),
\end{equation}
where $\widetilde{Z}^m_T$ is the logits produced by the teacher $m$ on the teaching data, and $\tau$ is the temperature.


\subsection{Student Model}
\textbf{Constructing Student Network}
The student model $S$ in our MTD framework has the same neural architecture as the teachers. It consists of an encoder to get the representation of the sentence and a softmax layer for prediction.

\noindent\textbf{Training Student Model}
The constructed student network $S$ is also trained on the teaching data $D_A$ with two objectives. One is to match the ground-truth hard labels. The other is to match the soft labels generated by multiple teachers.

Formally, we define the loss for  training the student network as follow:
\begin{equation}
\vspace{-0.1cm}
\label{distill_loss}
\small
\mathcal{L}_S= (1-\lambda)\mathcal{L}_{S,GT}+\lambda \mathcal{L}_{MKD},
\end{equation}
where $\mathcal{L}_{S,GT}$ denotes the ground-truth loss using one-hot hard labels. Note that the predictions for samples in the intersection set are also treated as ground-truth hard labels.  $\mathcal{L}_{MKD}$ denotes the knowledge distillation loss using multiple teachers' soft labels, $\lambda$ is the hyper-parameter to trade off $\mathcal{L}_{S,GT}$ and $\mathcal{L}_{KD}$.

The ground-truth loss $\mathcal{L}_{S,GT}$ is defined as the cross entropy ($CE$) between the student's predictions and the ground-truth labels.
\begin{equation}
\label{L_CE}
\small
\mathcal{L}_{S,GT} = \sum_{i=1}^{|\mathcal{D}_{A}|}CE(G(i),\widetilde{P}_S(i)),
\end{equation}
where $G(i)$ and $\widetilde{P}_S(i)$ denote the $i$-th element of the ground-truth labels and the student's predictions, respectively.

The knowledge distillation loss $\mathcal{L}_{MKD}$ is defined as the Kullback-Leibler divergence ($KL$) between  the student's predictions $\widetilde{P}_S$ and the soft labels by each of multiple (two) teachers.
\begin{equation}
\label{L_multiKD}
\small
\mathcal{L}_{MKD} = \sum_{m\in \{T1,T2\}}KL(\widetilde{P}^m_T||\widetilde{P}_S),
\end{equation}
where $\widetilde{P}^m_T$ denotes the probability distribution output by the teacher $m$.

\section{Experiments}
We conduct extensive experiments to verify the effectiveness of our MTD framework. We first introduce the experimental settings and then present the results and analysis.

\subsection{Dataset}
We evaluate our model on two public datasets: SemEval (SemEval-2010 Task 8)~\cite{semeval-2018} and TACRED (TAC Relation Extraction Dataset) \cite{prnn}.

(1) \textbf{SemEval} is a standard benchmark dataset containing  19 kinds of relations (including no\_relation). Note that the relations in SemEval are directed. For example, `Entity-Destination(e1,e2)' and `Entity-Destination(e2,e1)' are different relations.

(2) \textbf{TACRED} is a large-scale RE dataset which includes 41 undirected relation types such as ``\emph{per:age}'' and an extra ``\emph{no\_relation}''. It is typically used in the annual TAC knowledge base population competition.

The detailed data statistics along with the splits for SemEval and TACRED are shown in Table~\ref{dataset}.

\begin{table}[h]
\vspace{-1mm}
\caption{Statistics for two datasets.}
\label{dataset}
\setlength{\tabcolsep}{1mm}
\small
\vspace{-2mm}
\begin{center}
\begin{tabular}{cccccc}
\hline
\textbf{Dataset} & \textbf{\#Train} & \textbf{\#Dev} & \textbf{\#Test} & \textbf{\#Relations} & \textbf{\%No\_relation}\\
\hline
SemEval & 7,199 & 800 & 2,715 & 19 & 17.6\\
TACRED & 68,124 & 22,631 & 15,509 & 42 & 79.5\\
\hline
\end{tabular}
\end{center}
\vspace{-0.5cm}
\end{table}

\subsection{Compared Methods}
We compare our model with the following baselines.

(1) \textbf{PCNN} \cite{PCNN} utilizes a piecewise CNN with position embedding to model the word sequence.

(2) \textbf{PRNN} \cite{prnn} employs a bi-LSTM to encode the position, POS, NER, and token information. 

(3) \textbf{AGGCN} \cite{AGGCN} adopts GCN with multi-head attention to extract important information from the dependency tree. Note that AGGCN does not distinguish the relation direction.

(4) \textbf{NERO} \cite{nero} is a rule-based method. It first obtains candidate frequent patterns from the data and then employs expert knowledge to choose effective and correct patterns. It finally uses soft-matching for relation extraction. We use the manually specified patterns defined in NERO in this experiment.

(5) \textbf{KD4NRE} \cite{KD4NRE} utilizes knowledge distillation and statistical information to improve the performance. Note that its statistical information is extracted  on the basis of entire training set.

(6) \textbf{BERT${\rm _{EM}}$} \cite{bert4re} is a PLM based method with several variants. We adopt the BERT${\rm _{EM}}$ variant which performs the best with entity markers. 

(7) \textbf{Self-Training} \cite{self-training} is a SSRE method. It utilizes the prediction module to label the unlabeled data for  data augmentation, and it then iteratively trains on the augmented data. We train it two times with different seeds to get two models.

(8) \textbf{Mean-Teacher} \cite{mean-teacher} is a SSRE method. It makes use of the intersection set of predictions by two classifiers on unlabeled data and adds the top-ranked samples into the training set. 

(9) \textbf{Re-Ensemble} \cite{self-ensembling} is a SSRE method. It constructs two independent prediction modules and then uses the top-ranked  samples in the intersection set by two modules for  data augmentation.

(10) \textbf{DualRE} \cite{DualRE} is a SSRE method. It jointly trains a prediction module and a retrieval module to select the top-ranked  samples in the intersection set by two modules for  data augmentation. We use the point-wise variant as it performs better than the pair-wise one.

(11) \textbf{BERT w. gold labels} adopts the self-training method to train the BERT model, and it uses true labels to label the augmented data during each iteration. This baseline sets the upper bound for base SSRE methods. 

We re-implement four base SSRE methods with BERT${\rm _{EM}}$ as the encoder to keep pace with the state-of-the-art PLM based methods.

\subsection{Setup}
Following  existing SSRE methods \cite{DualRE,nero}, we sample 5\%, 10\%, and 30\% training data in SemEval, and 3\%, 10\%, and 15\% training data in TACRED as the labeled set, respectively. 50\% training data in SemEval and TACRED are sampled as unlabeled data whose labels are assumed unavailable for all models except the BERT w.gold labels method.
For all compared methods (supervised and semi-supervised), we follow the default hyper-parameter settings in original papers. For all SSRE methods, we select up to 10\% instances of the unlabeled data in the intersection set for data augmentation in each iteration and perform 10 iterations. The learning rate for the BERT${\rm _{EM}}$ encoder is set to ``5e-5''.

The parameters of our model are obtained on the development set. Specifically, the epoch for model training is set to 10, the batch size is 20, and the temperature of distillation is 2.4. The distillation loss weight $\lambda$ on SemEval and TACRED is set to 0.3 and 0.5, respectively.

We use $F_1$ as the main metric, and precision and recall as auxiliary metrics to evaluate the performance of all methods. The platform is a 24 GB NVIDIA RTX 3090 GPU.

\subsection{Main Results}
The baselines are classified into three categories. The first six methods (1)-(6) except NERO are supervised methods, where NERO is a rule-based one. The next four methods (7)-(10) are base SSRE methods. The last one (11) is also an SSRE method with the gold labels. We integrate the proposed MTD framework with all base SSRE methods.
The comparison results on SemEval and TACRED are shown in Table~\ref{SemEval} and Table~\ref{TACRED}, respectively.


It can be observed that our MTD framework can significantly enhance the performance of the base SSRE methods on both datasets and all ratios of labeled data. This clearly proves the effectiveness and generality of our proposed framework.
Moreover, we see that the trend is more obvious when the ratio is small. This is a very positive finding since we always wish to train a better classifier with less training data. We also notice that our MTD can improve the performance of the upper bound of SSRE methods, showing the effects of knowledge distillation.


Among the first six supervised methods, BERT${\rm _{EM}}$ is the best, with the remarkable enhancement over other supervised methods. This can be mainly due to the powerful expressive abilities of PLMs. The knowledge distillation method KD4NRE performs poorly when the proportion of labeled data is small, e.g., 5\% and 10\% on SemEval. The reason is that KD4NRE is designed for supervised learning and requires statistical information on the full data. Its performance on TACRED with 3\% labeled data is not that bad. This is because TACRED is much larger than SemEval. The performance of AGGCN on SemEval is extremely poor as it is based on the undirected connection graph while the relations in SemEval are directed. NERO is stable to different ratios of labeled data since it uses manually specified patterns.


\begin{table}[!htb]
\footnotesize
\caption{\label{SemEval} Comparison results on SemEval. All results are averaged over 5 runs with random seeds. ``$\ddagger$'' denotes the statistically significant improvements (i.e., two-sided t-test with $p<0.01$) over the corresponding base SSRE method.}
\vspace{-0.3cm}
\setlength{\tabcolsep}{1mm}
\begin{center}
\begin{tabular}{lccc}
\hline
  & 5\% ($F_1$) & 10\% ($F_1$) & 30\% ($F_1$) \\
\hline
(1) PCNN & $40.30_{\pm2.49}$ & $51.66_{\pm1.38}$ & $63.37_{\pm0.42}$\\
(2) PRNN & $54.66_{\pm0.89}$ & $62.49_{\pm0.59}$ & $69.14_{\pm1.02}$\\
(3) AGGCN & $5.99_{\pm1.26}$ & $6.34_{\pm0.86}$ & $6.88_{\pm2.17}$ \\
(4) NERO & $58.01_{\pm0.19}$ & $58.28_{\pm0.52}$ & $59.90_{\pm0.27}$ \\
(5) KD4NRE & $22.45_{\pm3.51}$ & $24.12_{\pm0.84}$ & $66.26_{\pm0.81}$\\
(6) BERT${\rm _{EM}}$ & $67.43_{\pm3.17}$ & $76.45_{\pm4.81}$ & $85.29_{\pm0.64}$\\
\hline
(7) Self-Training  & $73.61_{\pm2.22}$ & $80.48_{\pm0.73}$ & $85.47_{\pm0.37}$\\
~~~+ Our MTD  & $77.90^\ddagger_{\pm0.32}$ & $83.45^\ddagger_{\pm0.32}$ & $87.60^\ddagger_{\pm0.14}$\\
\hline
(8) Mean-Teacher & $73.20_{\pm0.87}$ & $80.85_{\pm0.27}$ & $85.66_{\pm0.33}$\\
~~~+ Our MTD & $76.47^\ddagger_{\pm0.87}$ & $83.66^\ddagger_{\pm0.18}$ & $87.40^\ddagger_{\pm0.20}$ \\
\hline
(9) RE-Ensemble  & $74.37_{\pm0.59}$ & $80.94_{\pm0.30}$ & $85.58_{\pm0.43}$\\
~~~+ Our MTD & $78.04^\ddagger_{\pm0.51}$ & $83.08^\ddagger_{\pm0.29}$ & $87.47^\ddagger_{\pm0.19}$\\
\hline
(10) DualRE & $75.15_{\pm1.32}$ & $81.14_{\pm0.84}$ & $85.97_{\pm0.56}$\\
~~~+ Our MTD & $78.08^\ddagger_{\pm0.64}$ & $84.06^\ddagger_{\pm0.14}$ & $87.62^\ddagger_{\pm0.31}$\\ \hline
(11) BERT w. gold labels & $85.45_{\pm0.69}$ & $86.21_{\pm0.25}$ & $86.72_{\pm0.33}$\\
~~~+ Our MTD & $88.04^\ddagger_{\pm0.17}$ & $88.20^\ddagger_{\pm0.25}$ & $88.68^\ddagger_{\pm0.45}$\\
\hline
\end{tabular}
\end{center}
\vspace{-0.3cm}
\end{table}

\begin{table}[!htb]
\footnotesize
\caption{\label{TACRED} Comparison results on TACRED. All results are averaged over 3 runs with random seeds. $\ddagger$'' denotes the statistically significant improvements (i.e., two-sided t-test with $p<0.01$) over the corresponding base SSRE method.}
\vspace{-0.3cm}
\setlength{\tabcolsep}{1mm}
\begin{center}
\begin{tabular}{lccc}
\hline
 & 3\% ($F_1$) & 10\% ($F_1$) & 15\% ($F_1$) \\
\hline
(1) PCNN & $45.39_{\pm0.78}$ & $50.42_{\pm1.00}$ & $52.25_{\pm0.67}$\\
(2) PRNN & $41.07_{\pm0.51}$ & $52.49_{\pm0.64}$ & $54.76_{\pm0.93}$\\
(3) AGGCN & $43.69_{\pm1.34}$ & $55.48_{\pm0.39}$ & $56.66_{\pm0.18}$ \\
(4) NERO & $45.97_{\pm0.41}$ & $47.07_{\pm0.61}$ & $47.48_{\pm0.92}$ \\
(5) KD4NRE & $43.52_{\pm0.55}$ & $55.59_{\pm0.41}$ & $58.07_{\pm0.31}$ \\
(6) BERT${\rm _{EM}}$ & $50.86_{\pm0.82}$ & $57.13_{\pm0.73}$ & $59.96_{\pm0.67}$\\
\hline
(7) Self-Training  & $52.68_{\pm0.19}$ & $58.96_{\pm0.14}$ & $61.43_{\pm0.40}$\\
~~~+ Our MTD  & $55.40^\ddagger_{\pm0.24}$ & $62.00^\ddagger_{\pm0.43}$ & $64.27^\ddagger_{\pm0.42}$\\
\hline
(8) Mean-Teacher & $52.20_{\pm0.45}$ & $59.60_{\pm0.19}$ & $61.47_{\pm0.75}$ \\
~~~+ Our MTD & $56.26^\ddagger_{\pm0.63}$ & $62.37^\ddagger_{\pm0.28}$ & $64.51^\ddagger_{\pm0.43}$\\
\hline
(9) RE-Ensemble  & $52.37_{\pm1.06}$ & $59.13_{\pm0.40}$ & $61.63_{\pm0.34}$\\
~~~+ Our MTD & $56.18^\ddagger_{\pm0.41}$ & $62.20^\ddagger_{\pm0.20}$ & $64.34^\ddagger_{\pm0.61}$ \\
\hline
(10) DualRE & $53.56_{\pm0.52}$ & $59.50_{\pm0.79}$ & $61.55_{\pm0.61}$\\
~~~+ Our MTD & $56.45^\ddagger_{\pm0.12}$ & $62.17^\ddagger_{\pm0.30}$ & $64.29^\ddagger_{\pm0.55}$\\ \hline
(11) BERT w. gold labels & $63.81_{\pm0.85}$ & $64.49_{\pm0.48}$ & $64.63_{\pm0.33}$\\
~~~+ Our MTD & $64.88_{\pm0.84}$ & $64.98_{\pm0.37}$ & $65.13_{\pm0.40}$\\
\hline
\end{tabular}
\end{center}
\vspace{-0.5cm}
\end{table}



\section{Analysis}
To get a deep insight into our proposed MTD framework, we conduct the ablation study, the complexity and loss landscape analysis, and the case study.

\subsection{Ablation Study}
We design a series of ablation studies, with 10\% labeled data on SemEval and TACRED, to examine the impacts of different components. All ablation studies are based on the simple self-training method.

\begin{table}[h]
\caption{Results for ablation study on SemEval. $\downarrow$ denotes a drop of F1 score.}
\label{Ablation1}
\vspace{-0.4cm}
\setlength{\tabcolsep}{1mm}
\small
\begin{center}
\begin{tabular}{cccc}
\hline
& \multicolumn{3}{c}{10\%SemEval}\\
 & precision & recall & F1 \\ \hline
Our MTD & $80.51_{\pm0.50}$ & $86.61_{\pm0.21}$  & $83.45_{\pm0.32}$ \\ \hline
Self-Training & $77.44_{\pm1.19}$ & $83.80_{\pm1.82}$  & $80.48_{\pm0.73}\downarrow$ \\
Self-Training\_I & $77.86_{\pm0.57}$ & $84.29_{\pm0.92}$  & $80.94_{\pm0.30}\downarrow$ \\
Intersection\_T & $78.33_{\pm0.52}$ & $85.88_{\pm0.22}$ & $81.93_{\pm0.35}\downarrow$ \\
Intersection\_S & $81.51_{\pm1.19}$ & $84.19_{\pm1.57}$  & $82.81_{\pm0.36}\downarrow$  \\
Distillation\_O & $80.14_{\pm0.99}$ & $83.51_{\pm1.48}$  & $81.78_{\pm0.51}\downarrow$ \\\hline
\end{tabular}
\end{center}
\vspace{-0.2cm}
\end{table}

\begin{table}[h]
\caption{Results for ablation study on TACRED. $\downarrow$ denotes a drop of F1 score.}
\label{Ablation2}
\vspace{-0.3cm}
\setlength{\tabcolsep}{1mm}
\small
\begin{center}
\begin{tabular}{cccc}
\hline
& \multicolumn{3}{c}{10\%TACRED} \\
 & precision & recall & F1 \\ \hline
Our MTD & $70.45_{\pm1.43}$  & $55.37_{\pm0.62}$ & $62.00_{\pm0.43}$ \\  \hline
Self-Training & $64.23_{\pm0.33}$  & $54.49_{\pm0.39}$ & $58.96_{\pm0.14}\downarrow$ \\
Self-Training\_I & $61.15_{\pm0.97}$  & $57.25_{\pm0.41}$ & $59.13_{\pm0.40}\downarrow$ \\
Intersection\_T & $63.79_{\pm3.34}$ & $57.83_{\pm3.27}$ & $60.55_{\pm0.68}\downarrow$\\
Intersection\_S & $64.31_{\pm1.45}$  & $57.72_{\pm0.53}$ & $60.83_{\pm0.48}\downarrow$ \\
Distillation\_O & $64.19_{\pm0.38}$  & $59.42_{\pm0.69}$ & $61.71_{\pm0.36}\downarrow$ \\\hline
\end{tabular}
\end{center}
\vspace{-0.4cm}
\end{table}

(1) \textbf{Self-Training} is the original SSRE method which contains only one model.

(2) \textbf{Self-Training\_I} is an extension of Self-Training, which runs Self-Training twice with different seeds and takes the samples in the intersection set for data augmentation during each iteration in semi-supervised learning.

(3) \textbf{Intersection\_T} trains two teacher models on the teaching data for one time and then applies the first teacher (its initial model is the same as that in Self-Training) to the test data, i.e., we remove the entire distillation from MTD.

(4) \textbf{Intersection\_S}  trains a separate student model on the teaching data without the distillation supervision.

(5) \textbf{Distilation\_O}  employs one teacher for distillation, i.e., we keep the partial distillation by treating one teacher's predictions as soft labels.

The results for ablation studies on SemEval and TACRED are shown in Table~\ref{Ablation1} and Table~\ref{Ablation2}, respectively.
We make the following notes for these results.

Firstly, all variants with reduced components cause performance drops. This proves that all components, including the multiple teachers, the separate student model, and the distillation technique, contribute to the entire framework.

Secondly, the influence of multiple teachers can be observed from ``Self-Training\_I'' and ``Distilation\_O'' . On one hand, ``Self-Training\_I'' raises the performance of ``Self-Training'' by using two models. On the other hand, ``Distilation\_O'' is worse than the complete MTD framework, showing that the knowledge distilled from one teacher is less effective than that from two teachers.

Thirdly, it is interesting to find that a separate student model ``Intersection\_I'' with random initialization performs better than the well-trained teacher model ``Intersection\_T''. This is probably caused by the overfitting issue and the biased distribution information learned by the teacher.

Finally, the effects of distillation can be confirmed by the superior performance  of  the MTD model and that of ``Distilation\_O'', which contain the complete and partial distillation component, and are the best and the second-best on TACRED.

We also see that ``Distilation\_O'' is a bit inferior to ``Intersection\_T'' and ``Intersection\_S'' on SemEval but is better on TACRED. This can be due to the property of two datasets. Taking a close look at the data, we  find that the proportion of intersection data to unlabeled data is 83.56\% in SemEval and 91.22\% in TACRED, while the ratio of negative (no\_relation) samples is 14.65\% in SemEval and 82.89\% in TACRED. This suggests that the intersection set in SemEval contains more high-quality samples.
In such a case, ``Intersection'' takes  a lead and the influence of ``Distilation\_O'' is not very obvious on SemEval. However, ``Distilation\_O'' still beats ``Self-Training\_I'' on this dataset. This infers the positive impacts of distillation from another point of view.

\begin{table}[h]
\caption{Complexity analysis. $h$ = $hour$, $M$ = $1\times10^6$.}
\vspace{-0.3cm}
\label{complex}
\setlength{\tabcolsep}{1mm}
\small
\begin{center}
\begin{tabular}{ccccc}
\hline
& \multicolumn{2}{c}{10\%SemEval} & \multicolumn{2}{c}{10\%TACRED}\\
 & time & space & time & space  \\ \hline
Self-Training & 0.41h & 124M & 3.74h & 124M\\
~~~+ Our MTD & 0.91h & 124M & 8.24h & 124M\\ \hline
Mean-Teacher & 0.98h & 249M & 10.15h & 251M\\
~~~+ Our MTD & 1.01h & 249M & 11.11h & 251M\\ \hline
RE-Ensemble & 0.97h & 249M & 10.47h & 250M\\
~~~+ Our MTD & 1.02h & 249M & 11.22h & 250M\\ \hline
DualRE & 0.98h & 249M & 10.66h & 250M\\
~~~+ Our MTD & 1.04h & 249M & 11.64h & 250M\\ \hline
\end{tabular}
\end{center}
\vspace{-0.5cm}
\end{table}

\begin{table*}[h]
\footnotesize
\caption{\label{casestudy}
Case study. The red and blue tokens denote the subject ($e_1$) and object ($e_2$) entity.
The top three predictions are presented in ascending order for teacher and student models, and the ground truth labels are underlined. }
\vspace{-0.2cm}
\setlength{\tabcolsep}{1mm}
\begin{center}
\begin{tabular}{l|lr|lr|lr}
\hline
\bf Instance (I) & \multicolumn{2}{c|}{\bf Teacher1 (T1) }  & \multicolumn{2}{c|}{\bf Teacher2 (T2)} &
\multicolumn{2}{c}{\bf Student (S)} \\

\hline
\tabincell{l}{I1: They include sending \\\textcolor{red}{e-mails} to remind customers \\about \textcolor{blue}{abandoned items}.} & \tabincell{l}{Content-Container(e2,e1) \\ \underline{Message-Topic(e1,e2)} \\ Entity-Origin(e1,e2)}   & \tabincell{r}{0.109 \\ 0.112 \\ 0.232} & \tabincell{l}{Instrument-Agency(e2,e1) \\ Entity-Origin(e1,e2) \\ \underline{Message-Topic(e1,e2)}} & \tabincell{r}{0.003 \\ 0.004 \\ 0.987}  & \tabincell{l}{Entity-Origin(e1,e2) \\ Entity-Destination(e1,e2) \\\underline{Message-Topic(e1,e2)}} & \tabincell{r}{0.026 \\ 0.051 \\ 0.786} \\

\hline
\hline
\tabincell{l}{I2: \textcolor{red}{Tree} \textcolor{blue}{roots} that grow on the\\ surface are difficult to mow or\\ walk over and can effect the \\growth and health of nearby \\grass and groundcovers.} & \tabincell{l}{Member-Collection(e2,e1) \\ Cause-Effect(e1,e2) \\ \underline{Component-Whole(e2,e1)}}   & \tabincell{r}{0.106 \\ 0.158 \\ 0.308} & \tabincell{l}{Member-Collection(e1,e2) \\ \underline{Component-Whole(e2,e1)} \\ Component-Whole(e1,e2)} & \tabincell{r}{0.170 \\ 0.233 \\ 0.374}  & \tabincell{l}{Cause-Effect(e1,e2) \\ Entity-Origin(e2,e1) \\  \underline{Component-Whole(e2,e1)}} & \tabincell{r}{0.032 \\ 0.113 \\ 0.644} \\
\hline
\hline

\tabincell{l}{I3: `` Partnering to Achieve Gre-\\ater Effectiveness in Preventing\\ Blindness,'' Kathy Spahn,\\ President and \textcolor{red}{CEO}, \textcolor{blue}{Helen}\\ \textcolor{blue}{Keller International}.} & \tabincell{l}{org:top\_mem-\\bers/employees \\ \underline{no\_relation} \\ per:title} & \tabincell{r}{~\\0.055 \\ 0.179 \\ 0.742}& \tabincell{l}{org:members \\ \underline{no\_relation} \\ org:founded\_by} & \tabincell{r}{0.006 \\ 0.448 \\ 0.530} & \tabincell{l}{per:title \\ org:top\_mem-\\bers/employees \\ \underline{no\_relation}} & \tabincell{r}{~\\0.029\\ 0.156 \\ 0.643} \\
\hline

\hline
\end{tabular}
\end{center}
\vspace{-0.5cm}
\end{table*}

\vspace{-0mm}
\subsection{Analysis on Computational Cost}
We show the results for complexity analysis in Table~\ref{complex}.
Since the student model in MTD is trained after the teacher model, we can see that the integration of MTD does not increase the space cost upon the base SSRE methods.  Furthermore, the introduction of our MTD model slightly increases the running time of Mean-Teacher, RE-Ensemble, and DualRE. Recall that all these SSRE methods have two models, and the main time cost is for the iterative semi-supervised learning while  MTD only needs to train the student model. Self-Training has the smallest  time cost because it only trains one model for semi-supervised learning.

\vspace{-1mm}
\subsection{Analysis on Loss Landscape}
\begin{figure}[h]
\vspace{-0.3cm}
\center{
\subfigure[Self-training]{
\includegraphics[scale=0.25]{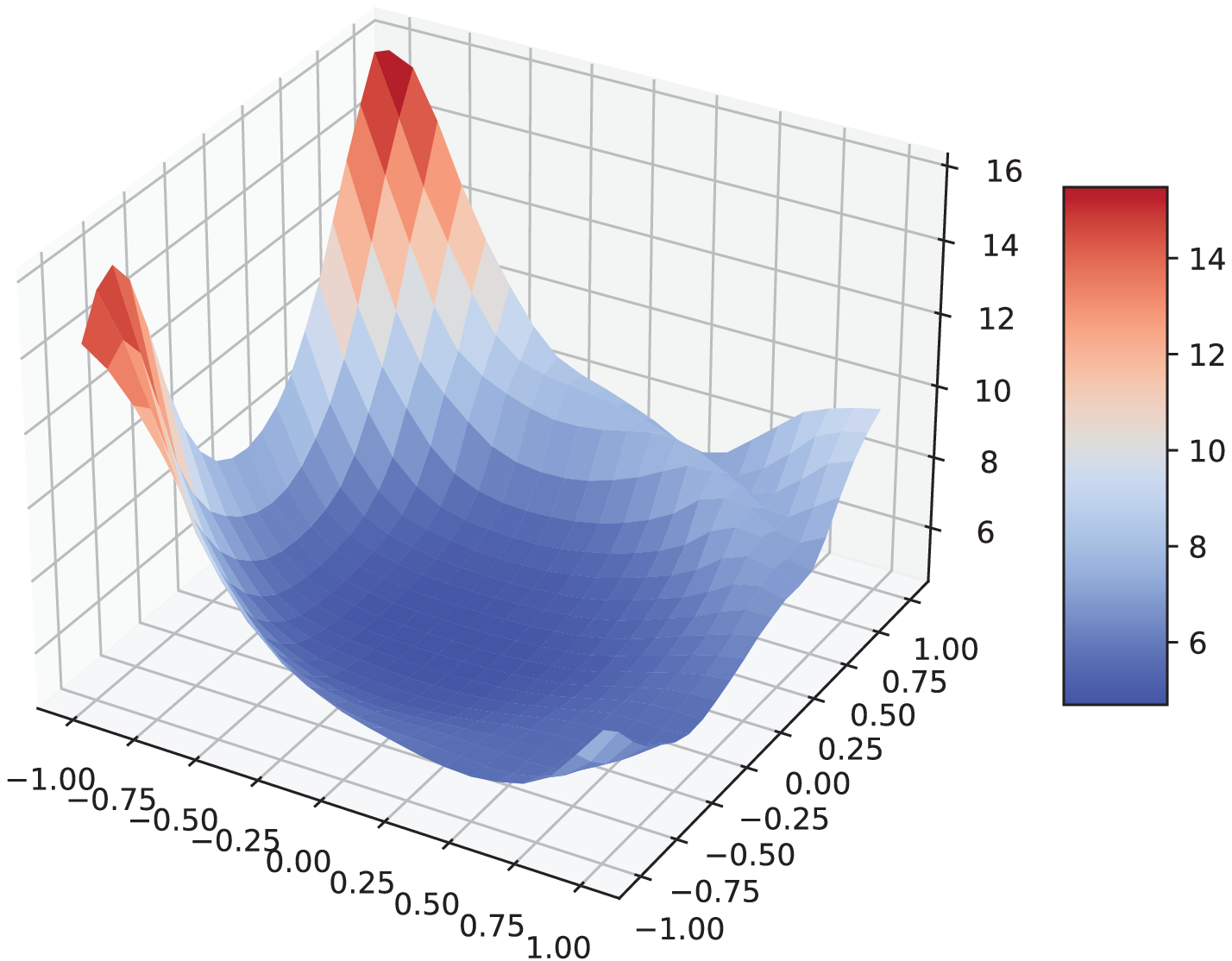}
}
\subfigure[Intersection\_S]{
\includegraphics[scale=0.25]{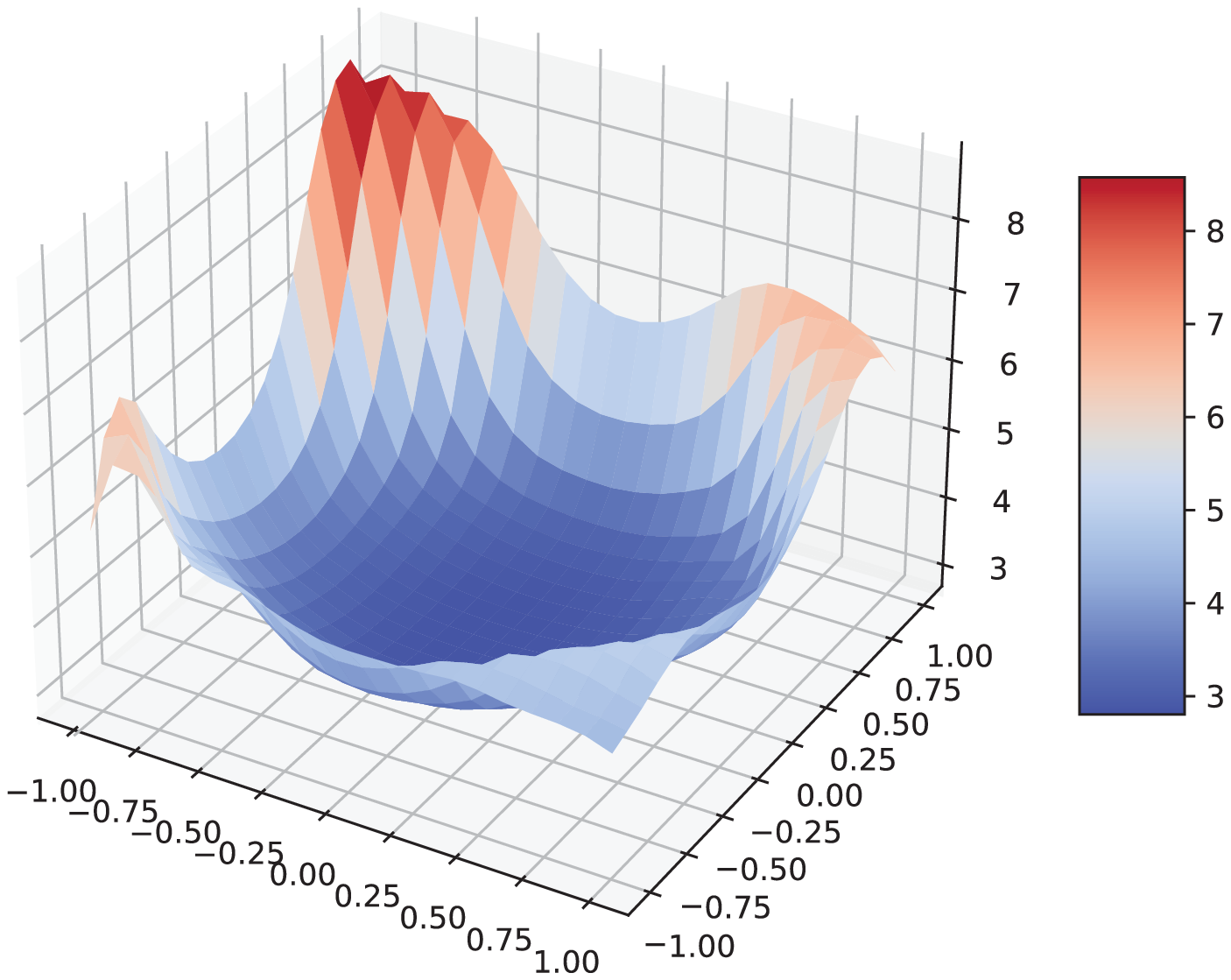}}
\subfigure[Distillation\_O]{
\includegraphics[scale=0.25]{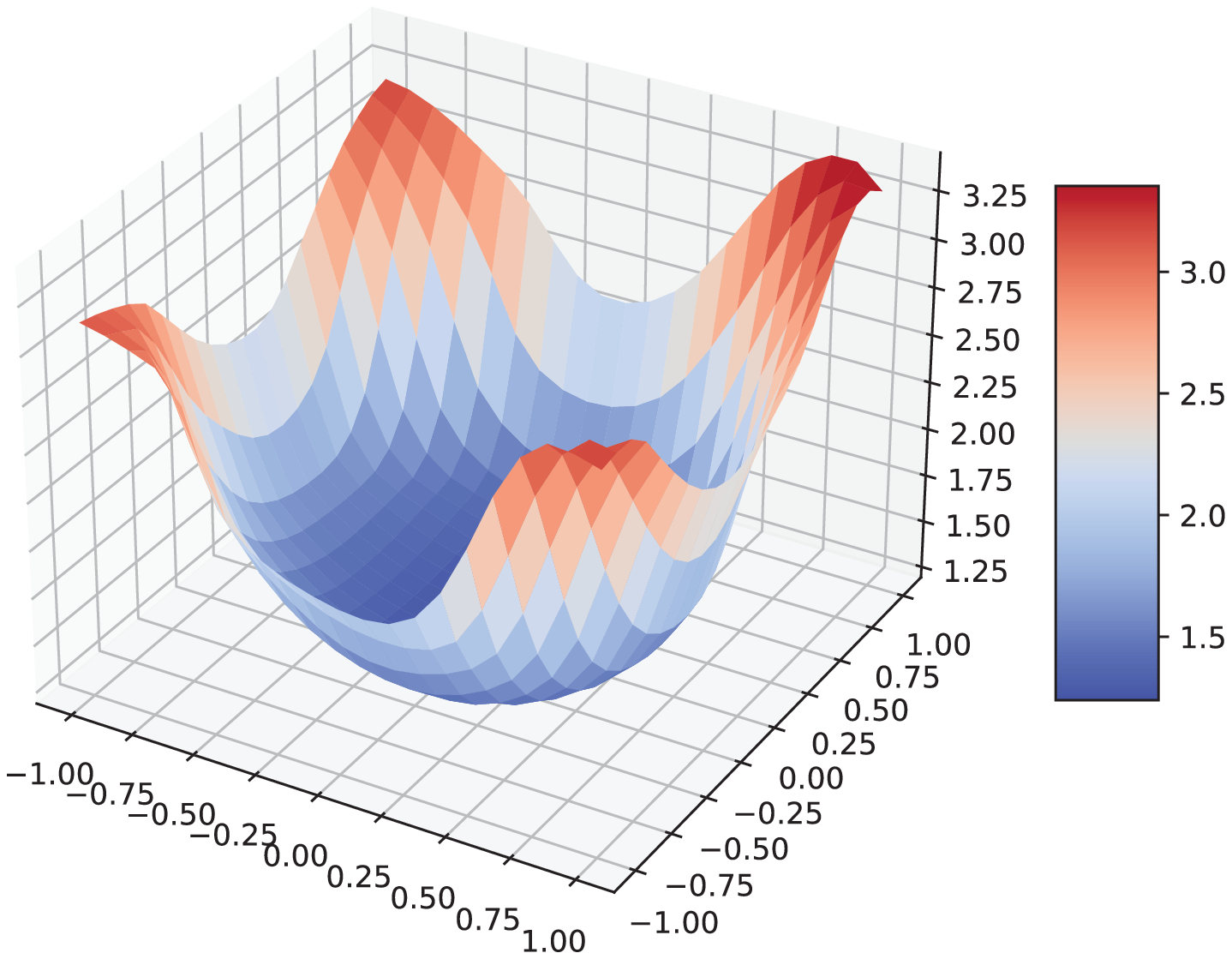}}
\subfigure[Self-training + Our MTD]{
\includegraphics[scale=0.25]{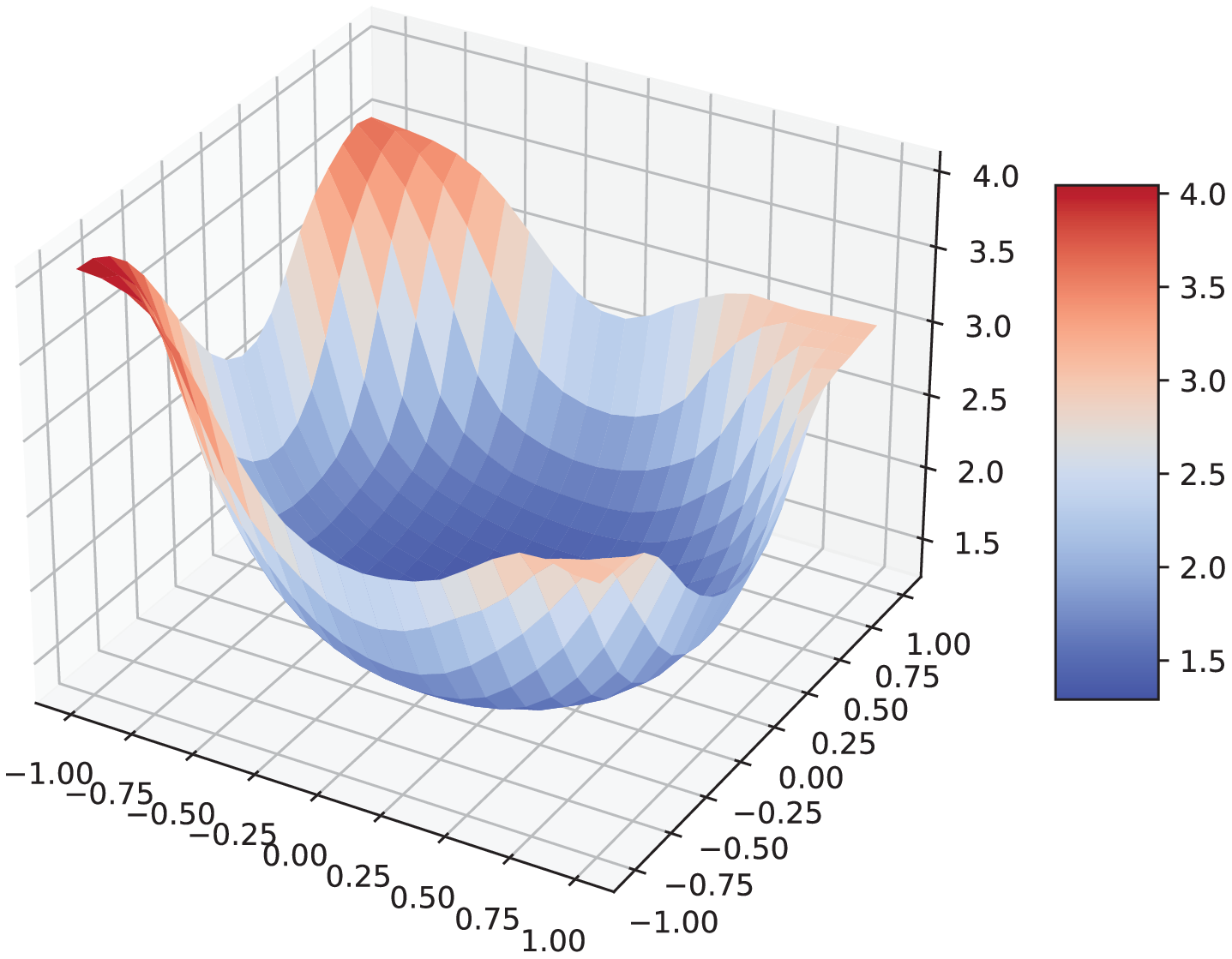}}}
\vspace{-0.3cm}
\caption{Visualization of the loss landscape.}
\vspace{-0.3cm}
\label{fig:vis}
\end{figure}

A recent research~\cite{visualize_loss2} shows that the loss landscape can support the analysis of KD methods.
We employ the state-of-the-art landscape visualization technique ~\cite{visualize_loss} to plot the loss surface of four representative methods in the ablation study. The results on SemEval are shown in Fig.~\ref{fig:vis}.

It is clear that our MTD produces the most flatter surface around the local minima. Distillation\_O has similar results. Meanwhile, the surfaces  for two methods without distillation (Self-training and Intersection\_S)  are much sharper.
The smooth loss landscape suggests that the distillation can result in higher accuracy and better generalization.

\subsection{Case Study}
Table~\ref{casestudy} shows several examples in SemEval and TACRED for a case study.

The first two instances I1 and I2 are from SemEval.
In I1, the teacher T1 assigns the wrong label ``Entity-Origin(e1,e2)'' with a low confidence. Meanwhile, T2 assigns the correct label ``Message-Topic(e1,e2)'' with a very high confidence.
In I2, both  T1 and T2 successfully identify the ``Component-Whole'' relation. However, T2 assigns a wrong direction and the totally correct prediction ranks the second.
In both instances, the student S can  make the correct prediction, indicating the effects of two teachers. If there is only one teacher, it might be hard for S to acquire the sufficient knowledge.

The last instance I3 is from TACRED. In I3, both teachers T1 and T2 make wrong yet different predictions. However, their second-highest scores are same and correct. The trained student network S inherits the knowledge distilled from teachers and further surpasses the teachers.



\section{Related Work}
Due to the space limitation, we only review the literature on closely related SSRE and KD methods in this section.

\subsection{Semi-Supervised Relation Extraction (SSRE)}
The  motivation behind SSRE is to reduce the manual efforts and to utilize the information in unlabeled data that is easy to obtain \cite{bootstrap_1,bootstrap_2,lp_1,active_1,active_2}. The main problem in SSRE is the semantic drift~\cite{semantic_drift07}.
The weak classifier learned on a small number of instances will inevitably cause the semantic drift problem~\cite{semantic_drift07}  in SSRE.
To address this problem, most SSRE methods adopt an ensemble strategy, where the simplest way is to select samples in  the intersection set of multiple classifiers. For example, Mean-Teacher~\cite{mean-teacher} employs different models with small perturbations on instances to make
consistent predictions. RE-Ensemble~\cite{self-ensembling} selects the samples to expand data based on the agreement of two prediction modules which are independently initialized. DualRE~\cite{DualRE} designs a retrieval module to assist the prediction module in generating more accurate annotations.

Though obtaining more reliable results, all existing SSRE methods discard the valuable difference set of predictions made by multiple classifiers. In view of this, we propose to make use of the class distribution information in the difference set and design a KD framework for this purpose.

\subsection{Knowledge Distillation (KD)}
KD is mainly for model compression~\cite{Bucilua_ModelCompression_kdd06,Ba_Deep_NIPS14,distillation_hinton,Fu_aaai21_kd4nlu}, where the knowledge extracted from a pre-trained teacher network provides guidance for a small student network.
KD has also been successfully applied to various fields, including computer vision~\cite{Yim_cvpr17_kd4vis,Zhang_cvpr18_kd4vis},
natural language understanding such as linguistic acceptability and textual
entailment~\cite{Clark:ACL19,Fu_aaai21_kd4nlu}, and recommender systems~\cite{Zhang_wsdm20_kd4rec,Kang_KDD21_KD4Rec}.

KD has not been considered by existing SSRE methods. We notice a seminal research~\cite{KD4NRE} distills the knowledge from the soft labels for  relation extraction, but it is designed for RE under the supervised scenario.
In contrast, we make the first attempt on exploiting KD for SSRE, where the task and the approach are completely different from those in ~\citet{KD4NRE}.



\section{Conclusion}
In this paper, we propose a novel model for SSRE. The key observation is that existing SSRE methods neglect the class distribution information in the difference set of  multiple models' predictions. Based on this observation, we design a multiple-teacher distillation (MTD) framework to transfer the distribution knowledge from teacher networks to the student network. MTD is simple and general, and it can be easily integrated into any SSRE method. MTD is also lightweight as it requires no extra space cost and introduces a small time cost. Extensive experiments on two popular datasets verify that our framework can significantly improve the performance of the base SSRE methods.

\bibliography{aaai22}
\end{document}